\pdfoutput=1

\documentclass[11pt]{article}

\usepackage[]{acl}

\usepackage{times}
\usepackage{latexsym}
\usepackage{graphicx}
\usepackage{epstopdf}
\usepackage{algorithm}
\usepackage{booktabs}
\usepackage{multirow}
\usepackage{array}
\usepackage{subfigure}
\usepackage{algpseudocode}
\usepackage{amsmath}
\usepackage{graphics}
\usepackage{epsfig}
\usepackage{color}
\usepackage{bm}
\usepackage{amsmath}
\usepackage{multirow}
\usepackage{graphicx}
\usepackage{array}
\usepackage{subfigure}
\usepackage{algpseudocode}
\usepackage{amsmath}
\usepackage{graphics}
\usepackage{epsfig}
\usepackage{booktabs}
\newcommand{\tabincell}[2]{\begin{tabular}{@{}#1@{}}#2\end{tabular}}
\usepackage{color}
\usepackage{bm}
\usepackage{amsfonts}
\usepackage{amsmath}
\usepackage{hyperref}
\usepackage[T1]{fontenc}

\usepackage[utf8]{inputenc}

\usepackage{microtype}

\title{Towards Better Chinese-centric Neural Machine Translation for Low-resource Languages}

\author{
	Bin Li$^{1}$\thanks{ \ \ Corresponding author.},
	Yixuan Weng$^{2}$,
	Fei Xia$^{2}$,
	Hanjun Deng$^{3}$\\[0.1cm]
	$^{1}$ College of Electrical and Information Engineering, Hunan University \\
	$^{2}$ National Laboratory of Pattern Recognition, Institute of Automation, Chinese Academy Sciences  \\
	$^{3}$ Experimental High School Affiliated to Beijing Normal University \\
	{libincn@hnu.edu.cn}, {xiafei2020@ia.ac.cn} \\
	\{{wengsyx, Hanjun\_Deng\}@gmail.com} \\
}

\begin{document}
\maketitle
\begin{abstract}
The last decade has witnessed enormous improvements in science and technology, stimulating the growing demand for economic and cultural exchanges in various countries. Building a neural machine translation (NMT) system has become an urgent trend, especially in the low-resource setting. However, recent work tends to study NMT systems for low-resource languages centered on English, while few works focus on low-resource NMT systems centered on other languages such as Chinese. To achieve this, the low-resource multilingual translation challenge of the 2021 iFLYTEK AI Developer Competition provides the Chinese-centric multilingual low-resource NMT tasks, where participants are required to build NMT systems based on the provided low-resource samples. In this paper, we present the winner competition system that leverages monolingual word embeddings data enhancement, bilingual curriculum learning, and contrastive re-ranking. In addition, a new Incomplete-Trust (In-trust) loss function is proposed to replace the traditional cross-entropy loss when training. The experimental results demonstrate that the implementation of these ideas leads better performance than other state-of-the-art methods. All the experimental codes are released at: \url{https://github.com/WENGSYX/Low-resource-text-translation}.
\end{abstract}
\section{Introduction}
Language, as the foundation of a nation and a symbol of culture, is the basis for the realization of the ``Belt and Road'' policy \cite{shisheng2019strategies}. It is of great significance to promoting economic and cultural exchanges. Recently, neural machine translation (NMT) \cite{kalchbrenner2013recurrent} mainly relies on the blessing of big data and computing power \cite{dabre2020survey}, and has reached its usefulness in rich resource scenarios such as Chinese and English \cite{zheng2021low}. However, the machine translation technology in the low-resource setting is still not mature enough \cite{zoph2016transfer}. For Chinese-centric NMT in low-resource, there are also problems such as sparse data with low-quality and complex forms, leading to the poor performance of modern NMT systems \cite{ijcai12021}.
\par In order to promote low-resource research on Chinese-centric multilingual translation and study the above problems, the 2021 iFLYTEK AI Developer Competition\footnote{http://challenge.xfyun.cn/topic/info?type=multi-langu-age-2021} supplies four Chinese-centric multilingual low-resource translation tasks, requiring participants to complete multilingual translation tasks with a few parallel data and a large amount of monolingual data as training samples. 
\par In this paper, we discuss some effective methods used in the champion NMT system for the multilingual translation of Chinese in low-resource. Specifically, we adopt the idea of dividing and conquering to solve the problems in the existing competition data set one by one. The monolingual corpus provided by the competition is utilized to enhance the bilingual data for the sparse training samples. As for the problem of poor quality of training corpus, we propose a new Incomplete-trust (In-trust) \cite{huang2021named} loss function replacing the traditional cross-entropy loss \cite{rubinstein2004cross} for better translation. In the training and post-processing stages, we adopted bilingual curriculum learning \cite{creese2004bilingual} and contrastive learning re-ranking \cite{khosla2020supervised} respectively to reduce the complexity of language translation in low-resource. Such methods can effectively improve the quality of the final results. Extensive experiments are conducted on different multilingual languages show that the implementation of our NMT system is competitive compared to other methods. In the end, we win the first place in the competition.
\par
The main contribution of this paper is the introduction and evaluation of the methods used in the Chinese-centric low-resource multilingual NMT competition, which can be mainly summarized as follows:
(1) we train and open source monolingual word vectors with the monolingual data set released by the competition for data enhancement of bilingual translation.
(2) We propose an In-trust loss function based on noise perception. Compared with the performance of the traditional cross-entropy loss, our method is competitive on two bilingual data sets.
(3) A variety of practical training and post-processing methods are used for better results. In the end, compared with other advanced methods, we achieve a significant improvement.
\section{Related works}
In this section, we describe the background on neural machine translation (NMT) and some common methods under low resources.
\par
\subsection{Neural machine translation}
The NMT \cite{kalchbrenner2013recurrent} model $\theta$ translates a sentence $\mathcal{X}$ from the source language to a sentence $\mathcal{Y}$ in the target language. With a parallel training corpus $\mathcal{S}=\left\{\left(\mathcal{X}^{i},\mathcal{Y}^{i}\right)\right\}_{i=1}^{N}$, NMT maximizes the log-likelihood of $y$ given $x$, assuming each $(x^{i}, y^{i})$ pair is independently and identically distributed:
$$\max _{\theta} \sum_{\left(\mathcal{X}^{i}, \mathcal{Y}^{i}\right) \in \mathcal{S}} \log p_{\theta}\left(\mathcal{Y}^{i} \mid \mathcal{X}^{i}\right).$$
The encoder-decoder structure is widely used in NMT, where the encoder converts the source sentence into a sequence of hidden representations and the decoder generates target words conditioned on the source hidden representations and previously generated target words. The encoder and decoder can be convolutional neural networks \cite{gehring2016convolutional}, recurrent neural networks \cite{dong2015multi}, and Transformer \cite{vaswani2017attention}. NMT has demonstrated effectiveness in the supervised learning setting, where labeled data is available and in this case is a parallel-corpus \cite{ranathunga2021neural}. The success of NMT methods has been reported in high-resource languages. However, paired sequences are usually expensive to collect, as it requires an expert to translate sequences $\mathcal{X}$ into another language $\mathcal{Y}$. 
\par
\subsection{The expansion of parallel corpus} In recent years, NMT has made some progress in low resources settings, especially low corpora \cite{ranathunga2021neural}. Many improvements are brought about by data-augmentation through synthetic parallel-corpora \cite{sennrich2015improving}. Transfer Learning approaches \cite{zoph2016transfer} have an orthogonal set of improvements in low-resource languages.
\par
\subsection{The use of other data} 
Due to the lack of parallel sentence pairs, leveraging data other than parallel sentences is essential in low-resource NMT \cite{wang2021survey}. In this paper, we categorize existing algorithms on low-resource NMT into two categories according to the data that is helpful. One is monolingual data. Using unlabeled texts to enhance the NMT models is a popular and effective method in various fields. Similarly, unlabeled monolingual data attracts a lot of attention because collecting monolingual data is much easier and cheaper than parallel data \cite{sennrich2015improving}. The other is the data from auxiliary languages. When training the NMT model, languages with similar syntax or semantics are both useful. Utilizing data from relevant and rich-resource languages has achieved great success in low-resource NMT. Recently, Neubig and Hu et al. consider choosing a resource-rich language in the same language family as an auxiliary language \cite{neubig2018rapid}, achieving significant improvements. Tan et al. \cite{tan2019multilingual} propose a language clustering method based on language embedding, which performs better than clustering by language family.
\begin{figure*}[t]
	\centering
	\includegraphics[scale=0.68]{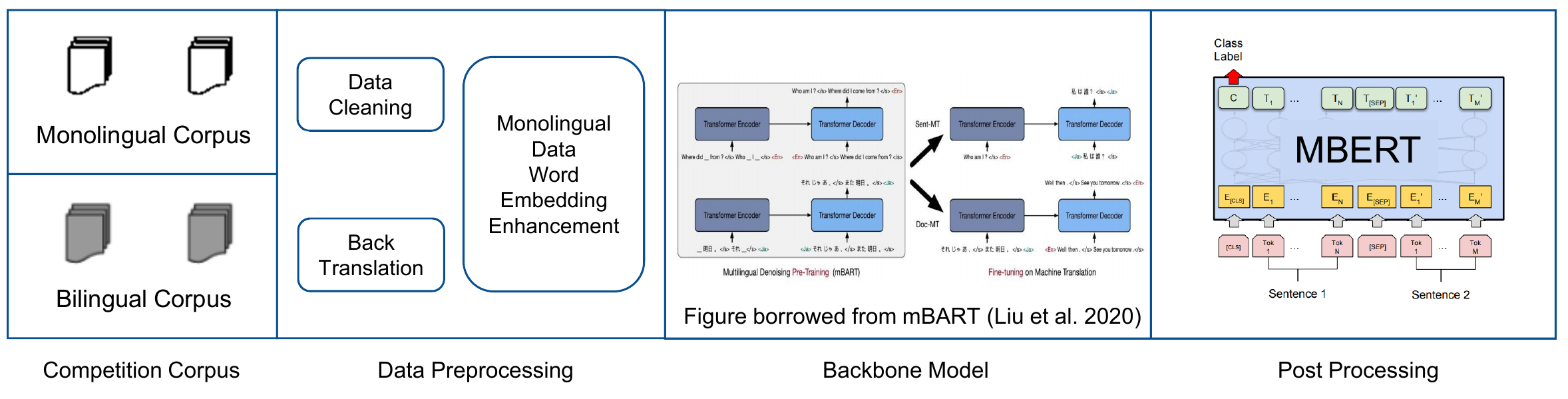}
	\caption{Architecture of the proposed NMT system.}
	\label{fig2}
	\vspace{-0.2cm}
\end{figure*}
\par
\subsection{Contrastive learning} 
In general, methods based on contrastive learning (CL) can well learn the differences between the observed data from other negative samples. CL has achieved good performance in both computer vision \cite{chopra2005learning, wang2015unsupervised} and natural language processing \cite{chen2020simple, liu2021simcls}. In the field of NMT, Yang et al. \cite{yang2019reducing} propose the method for leveraging CL for reducing word omission errors on NMT. Pan et al. \cite{mrasp2} apply CL for multilingual MT with the data augmentation for obtaining both the positive and negative training examples. Inspired by the work \cite{liu2021simcls}, we adopt CL in low-resource NMT with negative training examples being generated by the diverse beam search, thereby bridging the gap between the training and evaluation from various results.
\vspace{-0.1cm}
\section{Task introduction}
\subsection{Dataset}
The competition provides two low-resource multilingual translation corpora for Chinese, i.e., Ma-laysian and Indonesian. More specifically, the bilingual text translations include Chinese-Malaysian (Ch-Ma), Malaysian-Chinese (Ma-Ch), Chinese-Indonesia (Ch-In), and Indonesia-Chinese (In-Ch). This competition also provides 200,000 Malay-Chinese parallel text pairs and 200,000 Indonesian-Chinese parallel text pairs, 10,000,000 Chinese, 10,000,000 Malay, and 10,000,000 Indonesian monolingual texts. The development set is 1,000 sentences in each direction. The test set is 10,000 sentences in each direction.

\subsection{Evaluation metric}
This competition uses the automatic evaluation metric BLEU \cite{41}, for evaluation. As for the input text $S$ and the target text $T$. The calculation formula is as follows:
\begin{equation}
\small
BLEU=b({S}, T) \cdot \exp \left(\sum_{n=1}^{N} w_{n} \log P_{n}({S}, T)\right)
\end{equation}
where $N$ is 4 and the weighted parameter $ w_{n}$ is $1/4$, the percentile fraction is 1000. Furthermore, the N-gram can be calculated as follows:
\begin{equation}
\small
P_{n}({S}, T)=\frac{\sum_{k} \min \left(\text {Cnt}_{\text {clip}}(k,S ), \text {Cnt}_{\text {clip}}(k, T)\right)}{\sum_{k} \text {Cnt}(k, S)}
\label{bleu}
\end{equation}
where $k$ traverses all the n-grams candidates, and the $\text {Cnt}_{\text {clip}}(k, T)$ is the clipped n-grams. The weight $b(S, T)$ can be represented as follows:
\begin{equation}
b(S, T)=\begin{cases}
1\text{,}& \text { if }|T|>|S| \\
e^{(1-|S| / \mid T \mid)} & \text { if }|T| \leq|S|
\end{cases}
\label{bp}
\end{equation}
where $|T|$ represents the length of the target text, $|S|$ represents the length of the input text. 
\par
In addition, all the automatic evaluations are case-sensitive.
The character-based evaluation will be adopted for Malaysian-Chinese and Indonesian-Chinese translations. The Malaysian-Chinese and Indonesian-Chinese evaluations will convert the characters in the A3 area encoded by GB2312 from full-width to half-width. Chinese-Malaysian and Chinese-Indonesian translations adopt a word-based evaluation. The final ranking is based on the four unidirectional translation BLEU score averages.

\section{Method}
\subsection{System achitecture}
As shown in Figure \ref{fig2}, where the system is presented including competition corpus, data pre-processing, backbone model and post processing. The competition corpus includes the monolingual and bilingual data sets. The data pre-processing contains data cleaning and back-translation to obtain the high-quality data sets. The monolingual corpus is adopted for data enhancement with word embedding. We utilize the mBART model \cite{liu2020multilingual} as our baseline backbone model for multilingual NMT tasks in low-resource. The contrastive re-ranking is also used with MBERT \cite{sellam2021multiberts} in the post processing step for better results.
\subsection{Data preprocessing}
Since most of the competition data sets come from web crawling, and the submitted file is an extensible markup language (XML) type file that requires specific characters, we first deal with these special characters. The main operations include: 1) replace the \&amp with \&; 2) Replace \& in the target sentence with \&amp; 3. If the Traditional Chinese appears in the target sentence, it will be converted to Simplified Chinese.
\par
For the translation of bi-lingual data, we take into account its low-resource feature. This competition provides a large amount of monolingual data, so we adopt data enhancement. More precisely, we utilize data enhancement through back-translation \cite{sennrich2016improving}, where we adopt the Google API for providing well-formed translation. In addition, we use the monolingual corpus to train word vectors \cite{mikolov2013distributed} to replace source texts for data enhancement. As a  result, the original training data set is enhanced to 10 times the number of the original data. We conduct word vector training based on two monolingual corpora and open source to \url{https://github.com/WENGSYX/Malay-and-Indonesian-Word2vec}.
\subsection{Backbone baseline introduction}
We use the mBART model \cite{liu2020multilingual}  as our backbone to perform the NMT tasks, where the mBART is the NMT model performing seq2seq noise reduction auto-encoding pre-training on a large-scale monolingual corpus. We adopt the training bilingual text samples for fine-tuning the model. 
\subsection{In-trust loss function}
\begin{figure}[t]
	\centering
	\includegraphics[scale=0.6]{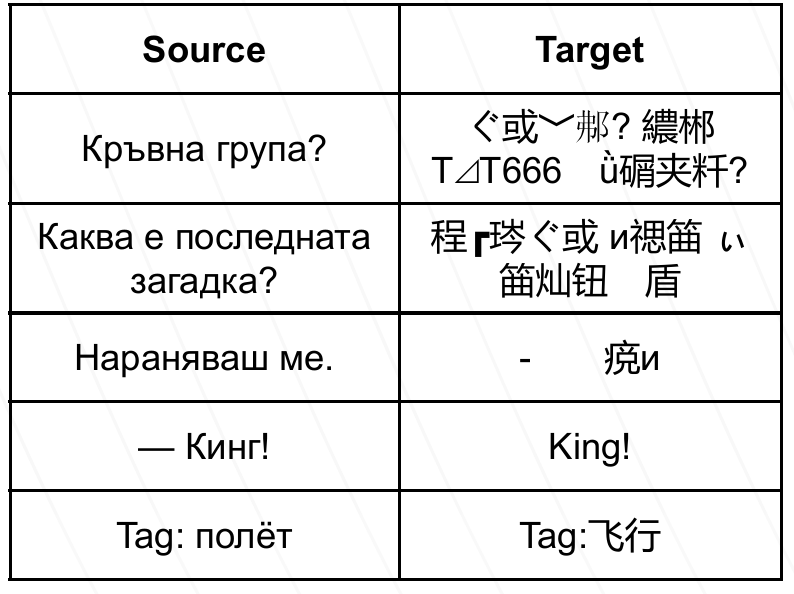}
	\caption{Noise samples in bilingual translation texts.}
	\label{fig3}
	\vspace{-0.4cm}
\end{figure}
As shown in Figure \ref{fig3}, the noise sample is cheery-pick from the competition datasets, where the noise font encoding hurts the model. Once over-fitting the noise samples, the translation performance of the model will be affected. The reasons may be the small scale and poor quality of the training set of small languages. Inspired by the work \cite{huang2021named}, the noise data sets may also provide knowledge. So we adopt the Incomplete-trust (In-trust) loss as the replacement for the original cross-entropy loss function, which is intended to train with uncertainty in the presence of noise. The new loss function is shown as follows
\begin{equation}
\begin{aligned}
&L_{DCE}=-p \log (\delta p+(1-\delta) q) \\
&L_{I n-t r u s t}=\alpha L_{C E}+\beta L_{D C E}
\end{aligned}
\end{equation}
where $L_{DCE}$ is an acceleration adjustment item, $p$ refers to the output information of the translation model, $q$ is the label, $\alpha, \beta$, and $\gamma$ are three hyper-parameters. The loss function also uses the label information and model output. These items form the $L_{DCE}$ item. $L_{I n-t r u s t}$ can effectively alleviate the model over-fitting noise sample.
\subsection{Contrastive re-ranking}
Re-ranking the translation results is considered to be a wise choice to increase the diversity of translation results. Hence, we adopt the contrastive re-ranking method for our NMT systems.
\par
\begin{figure}[t]
	\centering
	\includegraphics[scale=0.52]{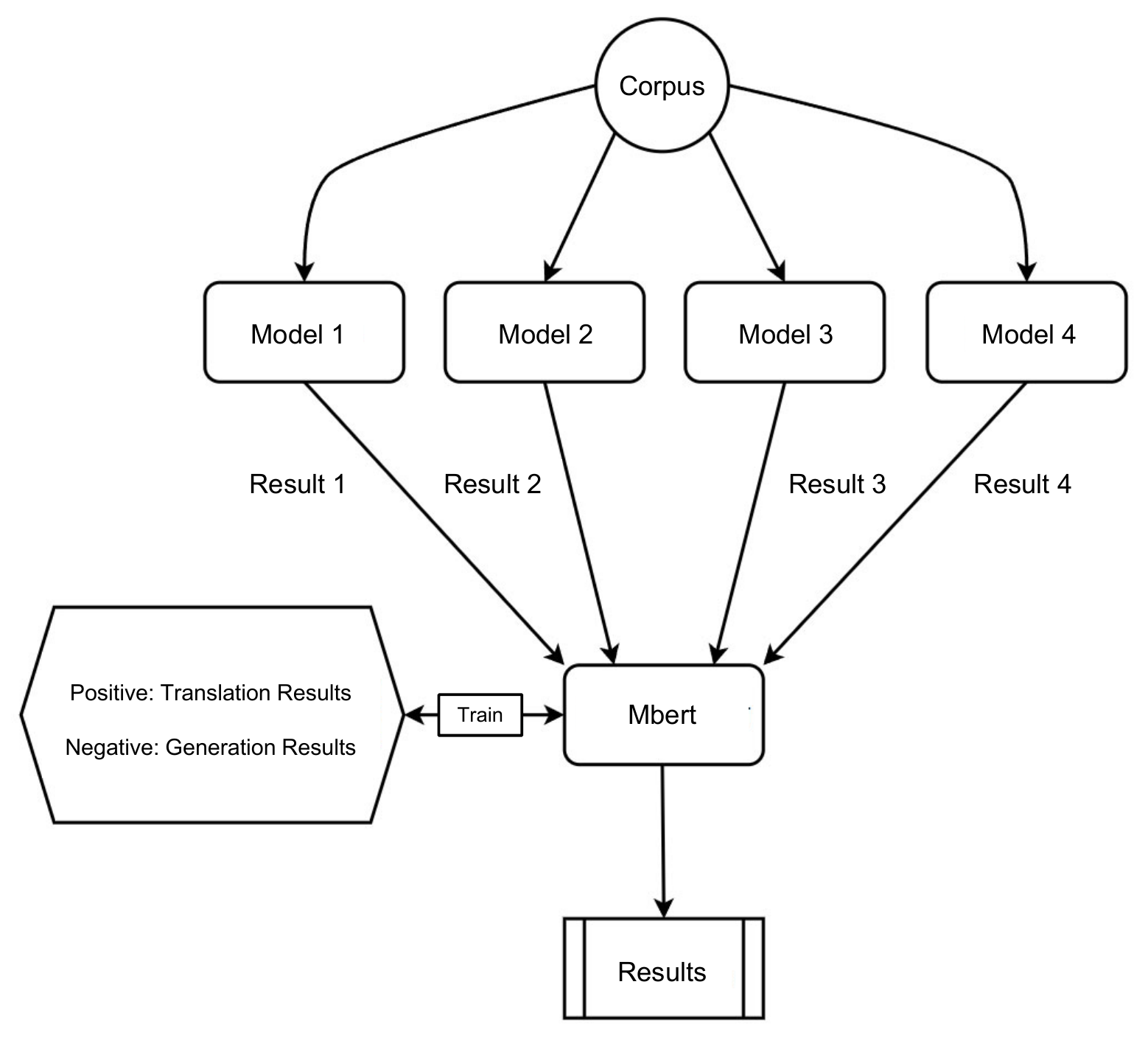}
	\caption{Overview of the contrastive re-reanking.}
	\label{fig45}
	\vspace{-0.3cm}
\end{figure}

As shown in Figure \ref{fig45}, where the re-ranking step is designed for better translation results from different models. After the decoded candidate results are obtained, we adopt the MBERT \cite{sellam2021multiberts} using the bilingual corpus to train the re-ranking model with contrastive learning. The positive samples are obtained from the bilingual corpus, while the negative samples are obtained from the Diverse Beam Search \cite{vijayakumar2016diverse}. Finally, we select the result with the highest confidence value of MBERT from multiple sets of translation candidate results.
\par
More specifically, $ h_{x}$ is the final hidden feature of the input source text. The representation of the contrastive target samples can be represented as $h_{T_j}$, $j \in [1,n]$. The $h_{T}^{+}$ represents the feature of the positive sample $T^{+}$, and $h_{T_j}^{-}$ is feature of negative sample $T^{-}$. 
\par 
A non-linear projection layer is added on top of the MBERT model for obtaining representation. The calculation of two types of the feature can be shown as follows:
\begin{equation}
\mathcal{L} = - \log \frac{e^{ \text{sim}\left( {h_{x}},  {h^{+}_{T}}\right) / \tau}}{\sum_{j=1}^{n} e^{ \text{sim}\left(h_{x}, h^{+}_{T} \right) / \tau} + e^{ \text{sim}\left(h_{x}, h^{-}_{T_j}\right) / \tau}}
\label{eq1}
\end{equation}
	\vspace{-0.2cm}
\section{Experiment}
\subsection{Compared baselines}
\begin{itemize}
	\item \textbf{mBART model} The mBART \cite{liu2020multilingual} adopts the BART model \cite{lewis2020bart} to perform on a large-scale monolingual corpus seq2seq denoising auto-encoding pre-training. The mBART can be fine-tuned via passing a variety of multilingual text of the language with a complete seq2seq module.
	\item \textbf{M2M model}
	The Many-to-many (M2M) multi-language translation model can directly translate between any 100 languages, and it also covers the competition corpus data. The model is composed of Transformer architecture \cite{vaswani2017attention}. After large-scale pre-training, different variants can be obtained. We use 1.2B model as strong baseline models for comparison.
	\item \textbf{Google translation engine} Google Translate\footnote{https://translate.google.com/} provide a strong and robust multilingual NMT system. We perform translation by calling open APIs, aiming to compare mature translation systems in a low-resource setting.
\end{itemize}
\subsection{Bilingual curriculum learning}
\begin{figure}[h]
	\centering
	\includegraphics[scale=0.5]{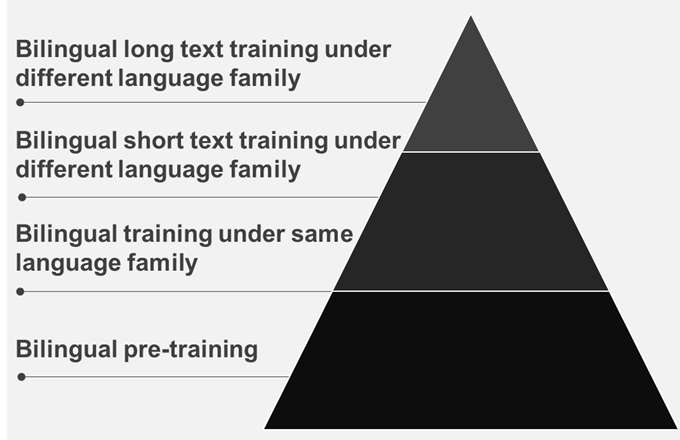}
	\caption{Overview of bilingual curriculum learning.}
	\label{fig4}
\end{figure}
According to the provided datasets by the competition, we adopt the bilingual
curriculum learning method, which is shown in the Figure \ref{fig4}. The training steps are from bottom to top, where the corpus used for the curriculum training needs to be carefully designed, so the shape is like a pyramid. By considering the language relationship between Indonesian and Malaysian and making reasonable use of the provided data, the following training steps are designed as follows in detail:
1).  Based on the pre-trained model, we use bilingual language training under the same language family, such as the language family of the Indonesian and Malaysian.
2). After obtaining the same language family NMT models, we then train the short text of bilingual corpus. It is because the model can learn a well-formed bilingual alignment in this step.
3). Finally, we concatenate the short source text sentence by sentence, and then splice them into a long text. It is because that the translation performance of the long text is relatively bad in low-resource.

\subsection{Implement details}
As for the mBART \cite{liu2020multilingual}, we adopt the large version following the default setting the same as the paper  {pan2021bert}, where the mBART\footnote{https://huggingface.co/facebook/mbart-large-50-many-to-many-mmt}  is adopted from the Transformers of the Huggingface \cite{wolf2019huggingface}. Note that the language of Malaysian is not covered in this model, so we will fine-tune the Malaysian-Chinese and Chinese-Malaysian for the downstream task. The AdamW optimizer is used with an initial learning rate of 1e-4 and annealed gradually after a warm-up epoch until it reached 1e-5. The weight hyper-parameter $\lambda$ is set to 0.5 to accelerate the whole training stage. Our fine-tuning stage is implemented with a batch size of 16 for 8 epochs.
\par
The M2M model \cite{fan2021beyond} can directly translate between the 9,900 directions of 100 languages, which covers the languages of this competition. This model is the strong baseline set in the zero-shot setting. We follow the walkthrough in the website\footnote{https://github.com/pytorch/fairseq/tree/main/examples/m2 \\ m\_100}, adopting 1.2B model for experimental comparison.
\par
The MBERT \cite{sellam2021multiberts} is utilized for contrastive re-ranking, The BERT-MT model is fine-tuned for 15 epochs with a batch size of 32. The positive and negative samples are set at the ratio of 1:4. We first train the mBART model for 8 epochs to generate the negative samples for feeding the MBERT model. The initial learning rate for the first encoder is 1e-5, and the others are 2e-4. The minimum learning rate is 1e-8 with the AdamW optimizer \cite{loshchilov2017decoupled}.
\begin{table}[]
	\centering
	\renewcommand\arraystretch{1.2}
	\setlength{\tabcolsep}{3mm}
	\caption{Translation results in different directions.}
	\begin{tabular}{ccc}
		\noalign{\hrule height 1pt}
		Direction               & Method                                      & BLEU                 \\
		\noalign{\hrule height 0.5pt}
		& mBART                                       & 26.77   \\
		
		& M2M                                                              &         27.72                                      \\
		
		& \begin{tabular}[c]{@{}c@{}}Google  Engine\end{tabular} &            27.41                                      \\
		
		\multirow{-4}{*}{Ch-Ma} & Ours                                                                &  \textbf{28.12}     \\				\noalign{\hrule height 0.5pt}
		& mBART                                                               &   22.17                      \\
		& M2M                                                                 &        23.14                                      \\
		& \begin{tabular}[c]{@{}c@{}}Google  Engine\end{tabular} &          {21.87}                                    \\
		\multirow{-4}{*}{Ma-Ch}                         & Ours                                                                &       \textbf{23.53}                                       \\
		\noalign{\hrule height 0.5pt}
		& mBART                                       &  27.42  \\
		
		& M2M                                                                 &          28.15                                     \\
		& \begin{tabular}[c]{@{}c@{}}Google  Engine\end{tabular} &              27.51                                \\
		\multirow{-4}{*}{Ch-In} & Ours                                                                &     \textbf{28.91}                                       \\
		\noalign{\hrule height 0.5pt}
		& mBART                                                               &    21.72                       \\
		& M2M                                                                 &         22.54                                      \\
		& \begin{tabular}[c]{@{}c@{}}Google  Engine\end{tabular} &              {22.43}                                 \\
		\multirow{-4}{*}{In-Ch}                         & Ours                                                                & \textbf{22.76}  \\     \noalign{\hrule height 1pt}                          
	\end{tabular}
	\label{onl}
\end{table}
\begin{table}[]
	\centering
	\renewcommand\arraystretch{1.2}
	\setlength{\tabcolsep}{2.4mm}
	\caption{Ablation study in final performance.}
	\begin{tabular}{cc}
		\noalign{\hrule height 1pt}
		Method     & BLEU-avg  \\ \noalign{\hrule height 0.5pt}
		mBART (CE loss)          & \textbf{24.52}         \\ 
		\tabincell{c}{+ Monolingual word embedding \\ data enhancement  }               & 24.91 \\ 
		+ In-trust loss             & 25.47     \\
		+ Bilingual curriculum learning                & 25.61      \\ 
		+ Contrastive re-ranking               & 25.83      \\ 
		\noalign{\hrule height 1pt}
	\end{tabular}
	\label{on3}
\end{table}
\begin{table}[]
	\centering
	\renewcommand\arraystretch{1.2}
	\setlength{\tabcolsep}{7.7mm}
	\caption{Online leaderboard.}
	\begin{tabular}{cc}
		\noalign{\hrule height 1pt}
		Method     & BLEU-avg  \\ \noalign{\hrule height 0.5pt}
		Ours          & \textbf{25.83}         \\ 
		Rank2                 & 25.70  \\ 
		Rank3             & 25.59     \\
		Rank4                  & 25.53      \\ 
		Rank5               & 25.42      \\ 
		Backbone baseline                & 24.52     \\ 
		\noalign{\hrule height 1pt}
	\end{tabular}
	\label{on2}
\end{table}
\section{Results}
It can be found that in the Figure \ref{onl} four low-resource multilingual translation tasks are provided by the competition, where the proposed NMT system surpasses other baselines. It shows that the proposed method is competitive with other baselines in low-resource translation scenarios for Chinese-centric. Further conclusions can be observed that in the translation direction of the Ch-Ma, the proposed system is 1.35 BLEU  higher than the mBART model, and the M2M model is higher than the result of the Google Translation Engine (Google Engine). At the same time, in the direction of Ma-Ch translation, the proposed system is 1.66 BLEU higher than the Google Engine. The reason may be that Google Translate has poor anti-noise ability in low-resource scenarios in Malaysian.  By using the noise perception loss function, the aligned text information of these translation samples can be effectively learned. In the translation direction of Ch-In, the proposed system is higher than the 1.49 BLEU score of the mBART model. In the translation direction of In-Ch, our system is 1.04 BLEU higher than the mBART model. These improvements show that the proposed method can effectively improve the quality of the final translation.
\par On the basis of mBART backbone baseline, adding monolingual data enhancement can effectively increase the 0.39 BLEU-avg score. It shows that monolingual data enhancement can effectively improve the translation performance of translation models, especially for low-resource scenarios. After replacing the In-trust loss function with the original cross-entropy (CE) loss, it has the largest improvement on BLEU-avg score, which is 0.56. This shows that with the noisy data sets, the proposed noise perception loss function (In-trust loss) can improve the robustness of the model compared with the original CE loss function. At the same time, we have adopted bilingual curriculum learning in the training process. It can effectively utilize the translation features of different languages in the same language family, and increase the BLEU-avg score of 0.14. Finally, combined with the contrastive re-ranking method, the final result performance increases by 0.22 BLEU value. The contrastive re-ranking method can effectively increase the diversity of results, and further improve the performance of the results.
\par 
In short, we finally rank first in the online rankings, as shown in the table \ref{on2}, which is sufficient to prove the effectiveness and practicality of our proposed system.
\section{Conclusion}
In this paper, we discuss the champion NMT system for the Chinese-centric multilingual low-resource translation tasks held by the 2021 iFLYTEK AI Developer Competition. Many baselines are implemented to compare with the proposed system, including the mBART model, M2M model and Google translation engine. We design the system with monolingual word embedding data enhancement, bilingual curriculum learning, and contrastive re-ranking. The results demonstrate that the proposed system outperforms all other baselines, achieving the best performance (top-1) in this competition. We have also open-sourced datasets, codes, and monolingual word vectors to promote the study of the Chinese-centric low-resource translation research. In future work, we will continue to open source more low-resource translation datasets on Chinese-centric. At the same time, we will also consider more efficient methods on multilingual low-resource datasets by building counterfactual texts.

\bibliographystyle{acl_natbib}
\bibliography{custom}


%
\end{document}